\documentclass[11pt,a4paper]{article}
\usepackage[hyperref]{acl2020}
\usepackage{times}
\usepackage{latexsym}

\usepackage{latexsym,graphicx}
\usepackage{amsmath,amssymb}
\usepackage{microtype}

\aclfinalcopy % Uncomment this line for the final submission
\title{Relation Extraction with Self-determined Graph Convolutional Networks}

\author{Sunil Kumar Sahu$^{1}$, Derek Thomas$^{2}$, Billy Chiu$^{1}$, Neha Sengupta$^{1}$, Mohammady Mahdy$^{1}$ \\
  $^{1}$Inception Institute of Artificial Intelligence, Abu Dhabi, United Arab Emirates \\
  $^{2}$PAX-AI, Abu Dhabi, United Arab Emirates \\
  \texttt{\{sunil.sahu,hon.chiu,neha.sengupta,mohammady.mahdy\}@inceptioniai.org } \\ 
  \texttt{derek.thomas@g42.ai } }
\date{}

\begin{document}
\maketitle
\begin{abstract}
Relation Extraction is a way of obtaining the semantic relationship between entities in text. The state-of-the-art methods use linguistic tools to build a graph for the text in which the entities appear and then a Graph Convolutional Network (GCN) is employed to encode the pre-built graphs. Although their performance is promising, the reliance on linguistic tools results in a non end-to-end process. In this work, we propose a novel model, the Self-determined Graph Convolutional Network (SGCN), which determines a weighted graph using a self-attention mechanism, rather using any linguistic tool. Then, the self-determined graph is encoded using a GCN. We test our model on the TACRED dataset and achieve the state-of-the-art result. Our experiments show that SGCN outperforms the traditional GCN, which uses dependency parsing tools to build the graph.
\end{abstract}

\section{Introduction}
\label{sec:intro}
Relation extraction (RE) aims at obtaining the semantic relationship between entities using text as a source of knowledge. For instance, from the text snippet, \textit{Steve Jobs and Wozniak co-founded Apple in 1976.}, we can infer that \textit{Steve Jobs} and \textit{Wozniak} have \textit{org:founded\_by} relation with \textit{Apple}. RE is an important subtask of information extraction that has significant applications in various higher-order NLP/IR tasks, such as question answering, knowledge graph completion and semantic search \cite{sunita-2008}. Earlier studies on RE were based on feature engineering. Such methods rely on linguistic and lexical tools to obtain the information required for such feature engineering \cite{zelenko2003}. Additionally, the performance of these methods is hindered by the sparse feature representation used by the models.

With the surge of neural networks, deep learning-based models have become prevalent. In these models, pre-trained word embeddings are employed to solve the feature sparsity problems. Deep learning based RE models can further be categorized along two lines: sequence-based and graph-based models. In sequence-based models, a word sequence is used to embed the text using convolution or recurrent neural networks \cite{Zeng14,zhou2016}. In graph-based models, the text is first converted into a graph using a dependency parser or other linguistic tools and then processed with a graph neural network which encodes neighborhood and feature information. Finally, the encoded graph features are used in RE. Along this line, \citet{liu2015dependency} and \citet{miwa-bansal2016} employed a bidirectional long short-term memory (BiLSTM) network and \citet{zhang2018graph} and \citet{wu-2019} employed a graph convolutional network (GCN) \cite{kipf2016semi} to encode the textual graph used in their work. Compared to sequence-based models, graph-based models have been shown to be effective in learning long-distance dependencies present in text \cite{zhang2018graph}.

Although the state-of-the-art results are obtained using graph-based models, they require external tools to build a graph for the text. Therefore, they are computationally expensive and not fully end-to-end trainable. While sequence-based models do not depend on external linguistic tools, they have been shown less effective for long text, especially when long-distance dependencies are required \cite{sunil2017}. To bridge this gap, we propose a Self-determined GCN (SGCN) which infers (self-determines) a graph for the text using a self-attention mechanism \cite{vaswani2017}, rather using any external linguistic tool. Then the self-determined graph is encoded using a GCN model. We evaluate the effectiveness of the SGCN on a RE task against several competitive baselines. In summary, our contributions are the following: 
\begin{itemize}
\item We build a novel graph-based model to encode text without the use of any linguistic tools. 
\item We show the effectiveness of the SGCN model on the RE task and achieve the state-of-the-art performance. 
\item We provide a comprehensive ablation analysis that highlights the importance of SGCN.
\end{itemize}

\section{Graph Convolutional Network (GCN)}
\label{sec:gcn}
 The GCN ~\cite{kipf2016semi} is an extension of a convolutional neural network, which encodes neighborhood information in a graph. Let \(G = (\mathbf{V},\mathbf{A},\textbf{X}) \) be a graph, where $\mathbf{V}$ represents the vertex set and $\mathbf{A} \in \mathbb{R}^{|\mathbf{V}|\times |\mathbf{V}|}$ typically represents a sparse adjacency matrix, where $\mathbf{A}_{(u,v)}$ = 1 indicates a connection from node $u$ to node $v$, else 0, and $\mathbf{X} \in \mathbb{R}^{|\mathbf{V}|\times d}$ represents node embeddings. Each GCN layer takes the node embedding from the previous layer and the adjacency matrix as input and outputs updated node representations. Mathematically, the new node embedding for node $v \in \mathbf{V}$ in the $l^{th}$ layer is:

\begin{equation}
\mathbf{z}_{v}^{(l+1)} = \sigma \left ( \sum_{u=1}^{n} A_{(u,v)} (\mathbf{W}^{(l)} \mathbf{z}_{u}^{(l)} + \mathbf{b}^{(l)} ) \right ),
\end{equation}

where $\mathbf{W}^{(l)} \in \mathbb{R}^{d \times o}$ and $\mathbf{b}^{(l)} \in \mathbb{R}^o$ are the parameters of the GCN at layer $l$ %, $\mathbf{z}_v^{(l+1)} \in \mathbb{R}^{o}$ will be the new node representation for $v$ 
and $\sigma$ represents a non-linear activation function.

\section{Self-determined Graph Convolution Network (SGCN)} 
\label{sec:sgcn}
As discussed in Section \ref{sec:intro}, most current works in NLP use a GCN to encode a pre-built graph, e.g., a dependency parsing graph \cite{Marcheggiani2017} or predicate-argument graph \cite{marcheggiani-bastings-titov:2018:NAACL}. Pre-built graphs require sophisticated tools that have been trained on manual annotations. Although such methods have demonstrated promising results in various NLP tasks, they are computationally expensive, not fully end-to-end trainable and not applicable to low-resource languages. To overcome these issues, our model dynamically self-determines multiple weighted graphs using a multi-head self-attention mechanism \cite{vaswani2017} and applies a separate GCN over each one.

\begin{figure}
    \centering
    \includegraphics[width=0.3\textwidth]{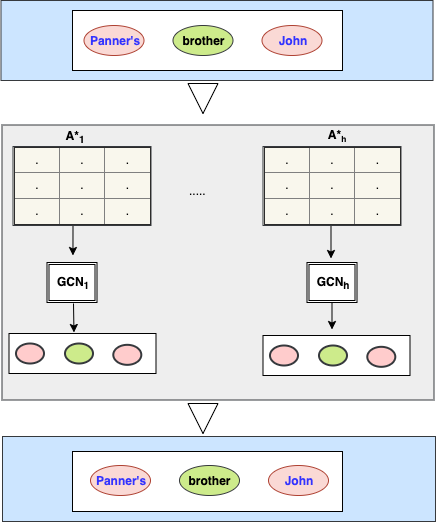}
    \caption{Model architecture of SGCN. First, $h$ adjacency matrices are self-determined from the text using a multi-head self-attention mechanism. Then, a separate GCN is employed for each graph to encode neighborhood information. Finally, the outputs of each GCN are concatenated. }
    \label{fig:sgcnn}
\end{figure}

Concretely, SGCN represents the words from the text as nodes in a graph and learns multiple adjacency matrices ($\mathbf{A}_1^*, \mathbf{A}_2^* \dots \mathbf{A}_h^*$),  $\mathbf{A}_i^* \in \mathbb{R}^{|\mathbf{V}|\times |\mathbf{V}|}$ in every layer of the GCN (as depicted in Figure \ref{fig:sgcnn}). Different from the $\mathbf{A}$ used in the traditional GCN, elements in $\mathbf{A}_i^*$ are not binary, but a mean normalized real numbers that represent the strength of the connection in the graph. Mathematically, for the $l^{th}$ layer, we compute the weight of the connection $u$ to $v$ for the $i^{th}$ head, $\mathbf{A}_{i^*_{(u,v)}}$, as:
 
% \begin{equation} 
% \label{eq1}
% A_i^*^{(l+1)}_{(u,v)} =  \footnotesize{\frac{1}{\mathbf{Z}^{(l)}_{v}} \mathbf{ReLU} \left ( \frac{\mathbf{K_i}^{(l)} \mathbf{z}^{(l)}_u . (\mathbf{Q_i}^{(l)} \mathbf{z}^{(l)}_v)^T}   {\sqrt{d} } \right )}, \\
% \end{equation}
\begin{equation} \label{eq1}
\begin{split}
\mathbf{M}_{i_{(u,v)}}^{(l+1)} &=  
\mathbf{ReLU} \left ( \frac{\mathbf{K_i}^{(l)} \mathbf{z}^{(l)}_u . (\mathbf{Q_i}^{(l)} \mathbf{z}^{(l)}_v)^T}   {\sqrt{d} } \right ) \\
\mathbf{A}_{i_{(u,v)}}^{*^{(l+1)}} &=  \frac{\mathbf{M}_{i_{(u,v)}}^{(l+1)}} {\sum_{u'\in V} \mathbf{M}_{i_{(u',v)}}^{(l+1)}} \\
\end{split}
\end{equation}
where $\mathbf{K_i}^{(l)}, \mathbf{Q_i}^{(l)} \in \mathbb{R}^{d \times d}$ are the trainable parameters. % and $\mathbf{Z}^{(l)}_v$ is the mean normalization value of node $v$. 
Once all $\mathbf{A_i}^*$s are obtained for the layer, we apply a GCN on each graph to encode the neighborhood information and concatenate the outputs. It is worth mentioning that the attention mechanism used in Eq. \ref{eq1} differs from the dot-product attention proposed by \citet{vaswani2017}. In this operation, we use the ReLU activation function \cite{relu2010} which can mask some of the attention weights by assigning them zero weight. This is more appropriate for the graph, since there are not always mutual connections between every node pair. In contrast to the traditional GCN, which uses the same connections in each layer, the SGCN determines the different connections.

\section{RE with SGCNs} 
\label{sec:re_sgcn}
For a given text $T = w_1, w_2 \cdots w_n$ and two target entities of interest $e_1$ and $e_2$ corresponding to the word (phrase) in $T$ , a RE model takes a triplet ($e_1$, $e_2$, $T$) as input and returns a relation for the pair, (including the no relation category) as output. 
%\footnote{We assume that both the target entities $e_1$, $e_2$, are provided with the task}. 
The set of relations used for inference are predefined. We first transform the text into a sequence of vectors using a pre-trained word embedding. Next, we employ a BiLSTM encoder to capture the context information in the vector sequence, which is then further used to represent the node of the graph. 

To further encode the long-distance context, we employ $k$-layer SGCNs in our model. As explained in Section \ref{sec:sgcn}, for each layer, SGCN dynamically determines the weighted connections for the graph using a self-attention mechanism and employs a GCN to propagate neighborhood information into nodes. Next, we employ a layer aggregation, originally proposed by \citet{xu2018representation}, in which all the SGCN layer outputs, along with a BiLSTM layer output, are concatenated and fed into a feed-forward layer. Finally, for relation classification, we follow \citet{zhang2018graph} and employ another feed-forward layer with a softmax operation on the concatenation of the sentence representation and both target entity representations. Sentence and entity representations are obtained by applying max-pooling over the entire sequence and average pooling to the position of  entities in the final representation, respectively. Following \citet{zhang2018graph} convention, now onward we refer to this model as \textbf{C-SGCN}. 

\section{Experiments and Dataset}
\label{Sec:exp}
We evaluate the performance of the C-SGCN model on the publicly available TACRED RE dataset \cite{tacred-2017}. It is the largest publicly available dataset for sentence-level RE. TACRED is manually annotated with $41$ categories of relations between subjects and objects. While the subject of these relations is PERSON and ORGANISATION, object consist of $16$ fine-grained entity types that include: DATE, LOCATION, TIME, etc. The dataset has $68124$, $22631$ and $15509$ instances for training, development and test out of which 79.5\% of the instances are labeled as no\_relation.

We employ the entity masking strategy to pre-process the dataset \cite{tacred-2017,zhang2018graph}, where each subject (and object similarly) will be replaced with a special $Subj-<NER>$ token. For instance, \textit{``MetLife$_{Obj}$ says it acquires AIG unit ALICO$_{Subj}$ for 15.5 billion dollars"} will become \textit{``Obj-Org says it acquires AIG unit Subj-Org for 15.5 billion dollars"}. Similar to other works \cite{tacred-2017,zhang2018graph,guo-etal-2019-attention}, we employed PoS tag embedding, and entity tag embedding\footnote{PoS and entity tags are provided with the dataset} along with word embedding to represent a word in the input of C-SGCN.

\subsection{Training and Hyper-parameter Settings}
In our model, ReLU activation function is employed in all GCN operation. We used $300$ dimension GloVe vector to initialize word embeddings and $30$ dimension random vectors to initialize PoS and entity tag embeddings. Parameters of the models are optimized using the stochastic gradient decent with batch size $50$ and initial learning rate of $0.3$. We used early stopping with patience equal to $5$ epochs in order to determine the best training epoch. %From the epoch $5$ we start to anneal the learning rate by a factor of $0.9$ every time the F1 score on the development set does not increase after an epoch. 
For other hyper-parameters, we perform a non-exhaustive hyper-parameter search based on the development set of the dataset. The dimension of SGCN and LSTM layer is set to $300$. We used $2$ layer SGCNs with $3$ heads in each for our experiments.  To prevent overfitting, we used dropout \cite{srivastava2014} in SGCN and LSTM layers with dropout rate equals to $0.5$. The remaining hyperparameter values are adopted from  \citet{zhang2018graph}.

\begin{table}[t!]
\centering
\scalebox{0.8}{
\begin{tabular}{l c c c } 
\hline
\textbf{Model} & \textbf{P} & \textbf{R} & \textbf{F1 (\%)}  \\
\hline
Patterns$\ast$ \cite{angeli2015}                  & 86.9 & 23.2 & 36.6\\
LR* \cite{surdeanu2012}                           & 73.5 & 49.9 & 59.4 \\
LR + Patterns$\ast$ \cite{angeli2015}             & 72.9 & 51.8 & \textbf{60.5} \\
\hline
SDP-LSTM$\ast$ \cite{xu-etal-2015-classifying}    & 66.3 & 52.7 & 58.7 \\
Tree-LSTM$\ast$ \cite{tai-etal-2015-improved}     & 66.0 & 59.2 & 62.4 \\
%GCN \cite{zhang2018graph}                   & 69.8 & 59.0 & 64.0 \\
C-GCN \cite{zhang2018graph}                  & 69.9 & 63.3 & 66.4 \\
S-GCN \cite{wu-2019}                         &   -  &  -   & 67.0 \\
%AGGCN \cite{guo-etal-2019-attention}        & 69.9 & 60.9 & 65.1 \\
C-AGGCN \cite{guo-etal-2019-attention}       & 73.1 & 64.2 & \textbf{68.2}\\
C-AGGCN$\dagger$ \cite{guo-etal-2019-attention}       & 69.6 & 66.0 & 67.8\\
%C-AGGCN$\bullet$                            &   -  &   -  & 67.7  \\
%LST-AGCN \cite{Sun2020}                     & - & - & 68.8 \\
\hline
%TRE \cite{Alt-2019}                         & 70.1 & 65.0 & 67.4 \\
%BERT-BiLSTM \cite{PengShi-2019}             & 73.3 & 63.1 & 67.8 \\
CNN$\ast$ \cite{kim-2014}                    & 75.6 & 47.5 & 58.3 \\
CNN-PE$\ast$ \cite{Zeng14}                   & 70.3 & 54.2 & 61.2 \\
LSTM$\ast$ \cite{zhang2015relation}          & 65.7 & 59.9 & 62.7 \\
PA-LSTM \cite{tacred-2017}                   & 65.7 & 64.5 & 65.1 \\
\textbf{C-SGCN-Softmax}                      & 69.3 & 65.4 & 67.3 \\
%\textbf{C-SGCN}                             & 70.2 & 66.2 & \textbf{68.2} \\
\textbf{C-SGCN}                              & 69.8 & 65.9 & \textbf{67.8} \\
\hline
\end{tabular}
}
\caption{Performance comparison of SGCN models against baselines. $\ast$ refers the performance was reported by \citet{tacred-2017} and $\dagger$ refers experiments conducted by the us on the shared code. The performances of the feature-based, sequence-based and graph-based model are separated in the first, second and third part of the table. The best F1 score in each section is highlighted. }
\label{tab:res_comp}
\end{table}

\subsection{Baseline Models}
We compare our C-SGCN model against several competitive baselines, which include feature engineering-based methods \cite{surdeanu2012,angeli2015}, sequence-based methods \cite{Zeng14,zhang2015relation,tacred-2017} and graph-based methods \cite{xu-etal-2015-classifying,tai-etal-2015-improved,zhang2018graph,wu-2019,guo-etal-2019-attention}. Apart from these, we additionally prepare \textit{C-SGCN-Softmax}, uses C-SGCN model with softmax to compute weighted graph in SGCN. To avoid any effects from the external enhancements, we don't consider methods that use BERT \cite{bert2019} or any other language model as pre-training in their models. We leave these experiments for future work.  

\subsection{Performance Comparison}
\label{Sec:perf_comp}
Table \ref{tab:res_comp} shows the performance comparison of SGCN models against all baselines. From the table, we can observe that C-SGCN outperforms all the feature-based and sequence-based models by a noticeable margin. Furthermore, compared to graph-based models, C-SGCN outperforms SDP-LSTM \cite{xu-etal-2015-classifying}, Tree-LSTM \cite{tai-etal-2015-improved}, C-GCN \cite{zhang2018graph} and S-GCN \cite{wu-2019}. However, C-SGCN's performance is same as C-AGGCN$\dagger$ \cite{guo-etal-2019-attention} in terms of F1 score. \footnote{With the code (\url{https://github.com/Cartus/AGGCN}) provided by the authors of C-AGGCN, $67.8$ is the best reproducible F1 score} 
It is worth mentioning that all of these works (except C-SGCN) employ a dependency parser to build the graph for the text. %, which is further encoded using a graph neural network. 
A dependency parsing requires external tool which is computationally expensive and time consuming.
In addition to this, our model C-SGCN performance is $0.5$ points higher than \textit{C-SGCN-Softmax} in terms of F1 score, verified the claim that use of ReLU activation function for computing edge weight is more appropriate for GCN framework. 

%In addition to this, among the methods that employ a pre-trained language model, SGCN outperforms TRE \cite{Alt-2019} and BERT-BiLSTM \cite{PengShi-2019}. In particular, the performance of BERT-C-SGCN is $0.8$ and $0.4$ points higher than that of TRE and BERT-BiLSTM in terms of F1 score, respectively. TRE employs Open AI GPT \cite{radford2018} on the transformer architecture of the RE model and BERT-BiLSTM, similar to our work, employs BERT in the BiLSTM based RE model. 

\subsection{Ablation Study}
We have demonstrated the strong empirical results obtained by the C-SGCN model. Next, we want to understand the contribution of each component employed in the model. We conduct an ablation test by removing some of these components. The ablated models are:
%(a) \textbf{No\_C-SGCN}, which represents the BERT-C-SGCN model without the C-SGCN component, i.e., the sentence and entity representations are obtained from the last layer output of the BERT model. 
(a) \textbf{No\_SGCN} is the C-SGCN model without the SGCN component, i.e., the sentence representation and entity representations are obtained from the output of the BiLSTM layer.
(b) \textbf{No\_LSTM}, represents the C-SGCN model without BiLSTM. Finally, 
(c) \textbf{No\_LA},  represents the C-SGCN model without the layer aggregation, i.e., the last layer output of SGCN is used to obtain the sentence and entity representations.

\begin{table}
\centering
\scalebox{0.7}{
\begin{tabular}{l c c c } 
\hline
\textbf{Model} & \textbf{P} & \textbf{R} & \textbf{F1 (\%)}  \\
\hline
%\textbf{C-SGCN}                         & 70.2 & 66.2 & \textbf{68.2} \\
\textbf{C-SGCN}                          & 69.8 & 65.9 & \textbf{67.8} \\
\textbf{No\_SGCN}                        & 71.7 & 61.6 & 66.3 \\   
\textbf{No\_LSTM}                        & 68.5 & 43.9 & 53.5 \\   
\textbf{No\_LA }                         & 75.2 & 55.9 & 64.1 \\   
\hline
\end{tabular}
}
\caption{Ablation analysis on the test set of TACRED dataset.}
\label{tab:ablation}
\end{table}

\begin{figure} 
    \centering
    \includegraphics[width=0.6\textwidth]{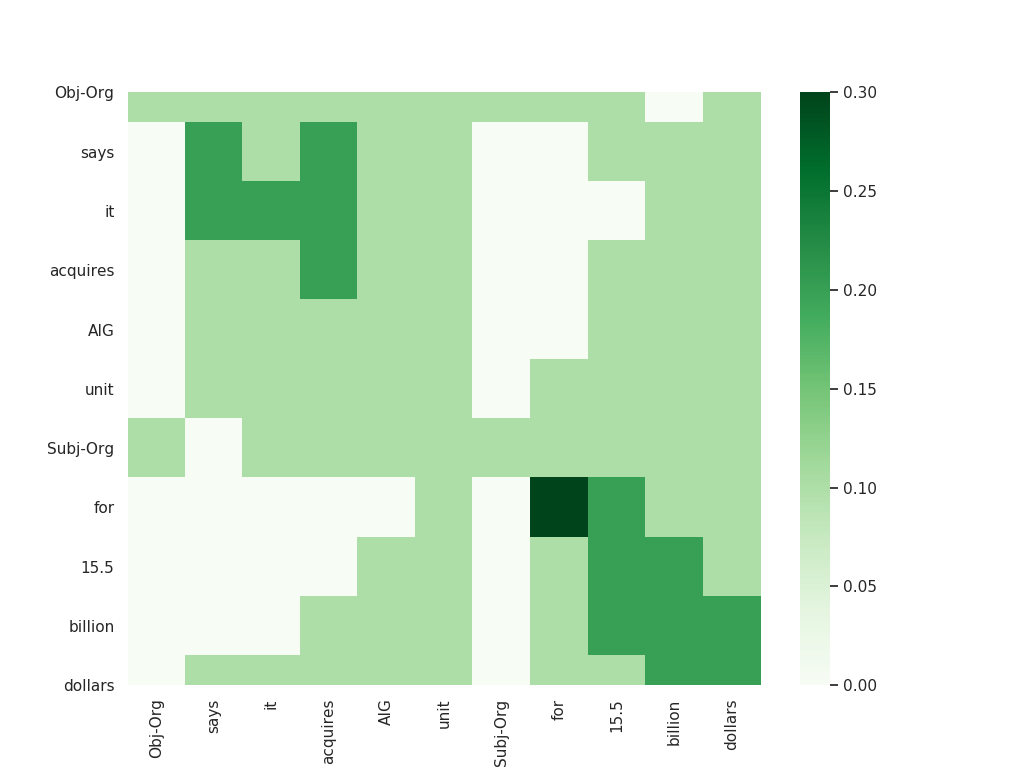}
    \caption{Heat map of the adjacency matrix determined by SGCN in first layer. The sentence is 'Obj-Org says it acquires AIG unit Subj-Org for 15.5 billion dollars'. Heat map values are approximated to one decimal point.}
    \label{fig:heat}
\end{figure}

Table \ref{tab:ablation} shows the results of the ablation study performance. From the table, we can observe that all the components employed in C-SGCN models have a noticeable contribution to the overall all performance. In particular, the performance of No\_SGCN is $1.5$ points lower than the C-SGCN in terms of F1 score, demonstrating the strong contribution of SGCN.

\subsection{Interpretation of the Self-determined Graph}
 The proposed approach dynamically determines the weighted graph for the text using a self-attention mechanism. In this section, we try to visualise the graph by plotting a heat map of the adjacency matrix obtained by the model. We wish to examine whether the SGCN indeed learns the connections that are important for the relation extraction task. Figure \ref{fig:heat} depicts the heat map figure of sentence \textit{``Obj-Org says it acquires AIG unit Subj-Org for 15.5 billion dollars"}. The sentence expresses the \textit{org:parents} relation between subject and object. From the figure, one can observe that the SGCN can infer a connection from the target entities to most of the other words in the text. In addition to this, it also inferring a strong self-connection weight to the words that are important for the prediction of the relation. Finally, the connections in the inferred graph are not symmetric.

\section{Related Work}
\label{rel_work}
RE is a well-studied field of knowledge extraction. Traditional feature-based methods rely on manual features obtained from various tools and lexical resources \cite{culotta2004}. Recently, various neural network based methods have also been applied for RE tasks. These include convolutional neural networks \cite{Zeng14}, recurrent neural networks \cite{zhou2016}, transformer networks \cite{patrick2018} and graph-based networks, e.g., Graph LSTMs \cite{peng2017} and GCNs \cite{zhang2018graph,wu-2019,guo-etal-2019-attention}. \citet{wu-2019} and \citet{zhang2018graph} employed a GCN to encode the dependency graph of the text. In their works, dependency graph is obtained using linguistic tools. \citet{guo-etal-2019-attention} employed dependency graph to initialize the first block of their GCN model and the attention guided self-attention layer is included in subsequent blocks. In \citet{guo-etal-2019-attention}, the idea behind self-attention layer is to dynamically prune or underweight the unimportant edges in the graph using soft-attention. However, we employ a self-attention layer to obtain a graph which will further be encoded using GCN.
%However, we employ a GCN to encode a self-determined graph, obtained automatically using a self-attention mechanism.

GCNs have been studied in various domains, using a variety of graphs, e.g., social network graphs \cite{kipf2016semi}, chemical reaction network graphs \cite{Coley2018} etc. In text, GCNs are employed to encode non-local dependencies present between the words in a text. They have been successfully used in co-occurrence graphs \cite{yao2019graph}, predicate-argument graphs \cite{marcheggiani-bastings-titov:2018:NAACL}, dependency parsing graphs \cite{zhang2018graph} and heterogeneous graph \cite{sunil2019}. To the best of our knowledge, this is the first work that employs a GCN on a fully self-determined graph.

 \section{Conclusion and Future Works}
 \label{sec:conc}
In this work, we proposed a novel model, C-SGCN, for the RE task. Our model dynamically determines the graph for the text using a self-attention mechanism. Although the proposed model is evaluated on the RE task, it is generic and can be applied for other tasks. 
%The experimental results demonstrate the effectiveness of our proposed SGCN. 
Experimental results show that our model achieves comparable performance to the state-of-the-art neural models that uses dependency parsing tool to obtain a graph for the text. 

Recently, several studies have demonstrated that employment of a pre-trained language model in end-to-end neural models further improves the performance of the downstream task, in future, we will try to incorporate a pre-trained language model with proposed C-SGCN model and improve the performance of RE. Besides, the applicability of SGCN in other text mining tasks are yet to be validated.

%\newpage 
\bibliography{acl2020}
\bibliographystyle{acl_natbib}

\end{document}